\documentclass[preprints,article,accept,moreauthors,pdftex]{Definitions/mdpi} 

\firstpage{1} 
\pubvolume{0}
\issuenum{0}
\articlenumber{0}
\pubyear{2022}
\copyrightyear{2022}
\hreflink{https://doi.org/}  
\pdfoutput=1

\usepackage{graphicx}
\usepackage{etoolbox}

\Title{FIgLib \& SmokeyNet: Dataset and Deep Learning Model for Real-Time Wildland Fire Smoke Detection}

\TitleCitation{FIgLib \& SmokeyNet: Dataset and Deep Learning Model for Real-Time Wildland Fire Smoke Detection}

\Author{Anshuman Dewangan $^{1}$, Yash Pande $^{1}$, Hans-Werner Braun $^{2}$, Frank Vernon $^{3}$, Ismael Perez $^{2}$, Ilkay Altintas $^{2}$, Garrison W. Cottrell $^{1}$ and Mai H. Nguyen  $^{2,}$*} 

\AuthorNames{Anshuman Dewangan, Yash Pande, Hans-Werner Braun, Frank Vernon, Ismael Perez, Ilkay Altintas, Garrison W. Cottrell and Mai H. Nguyen}

\AuthorCitation{Dewangan, A.; Pande, Y.; Braun, H.-W.; Vernon, F.; Perez, I.; Altintas, I.; Cottrell, G.W.; Nguyen, M.H.}

\address{%
$^{1}$ \quad Computer Science and Engineering Department, University of California, San Diego,  CA 92093, USA; adewangan@ucsd.edu (A.D.); ypande@ucsd.edu~(Y.P.); gary@ucsd.edu (G.W.C.)\\
$^{2}$ \quad San Diego Supercomputer Center, University of California, San Diego, CA 92093, USA; \linebreak hwb@ucsd.edu (H.-W.B.); i3perez@sdsc.edu (I.P.); ialtintas@ucsd.edu (I.A.)\\
$^{3}$ \quad Scripps Institution of Oceanography, University of California, San Diego,  CA 92093, USA; flvernon@ucsd.edu} 

\corres{Correspondence: mhnguyen@ucsd.edu}

\abstract{The size and frequency of wildland fires in the western United States have dramatically increased in recent years. On high-fire-risk days, a small fire ignition can rapidly grow and become out of control. Early detection of fire ignitions from initial smoke can assist the response to such fires before they become difficult to manage. Past deep learning approaches for wildfire smoke detection have suffered from small or unreliable datasets that make it difficult to extrapolate performance to real-world scenarios. In this work, we present the Fire Ignition Library (FIgLib), a publicly available dataset of nearly 25,000 labeled wildfire smoke images as seen from fixed-view cameras deployed in Southern California. We also introduce SmokeyNet, a novel deep learning architecture using spatiotemporal information from camera imagery for real-time wildfire smoke detection. When trained on the FIgLib dataset, SmokeyNet outperforms comparable baselines and rivals human performance. We hope that the availability of the FIgLib dataset and the SmokeyNet architecture will inspire further research into deep learning methods for wildfire smoke detection, leading to automated notification systems that reduce the time to wildfire response.}


\begin{document}
\section{Introduction}

Climate change has had a devastating impact on California in the form of increased wildfire activity and intensity. In~2018 alone, 8527 fires burned an area of 1.9 million acres in California (7700 km$^2$; nearly 2\% of the state’s area), with~an estimated economic cost of USD 148.5 billion~\cite{wang2021economic}. It is therefore imperative to detect and react to fire ignitions before they grow out of control. Currently, fire management teams rely on a combination of fire lookout experts, human-monitored camera feeds, and~public reports to detect fire ignitions. However, it can take much longer than the first few crucial minutes for a fire to be reported using these existing methods, especially in areas with less human activity. Automated techniques using computer vision and deep learning hold promise in addressing this need. Deep learning-based wildfire smoke detection systems can accurately and consistently detect wildfires and provide valuable intel to reduce the time to alert~authorities.

The goal of a wildfire smoke detection system can be structured as a binary image classification problem to determine the presence of smoke within a sequence of images. Priorities include quick time-to-detection, high recall to avoid missing potential fires, high precision to avoid frequent alarms that undermine trust in the system~\cite{govil2020preliminary}, and~efficient performance to operate in real time on edge devices. However, the~task proves challenging in real-world scenarios given the transparent and amorphous nature of smoke; faint, small, or~dissipating smoke plumes; and false positives from clouds, fog, and~haze. While the idea of an automated wildfire smoke detection system has been previously explored, the~difficulty of acquiring a large, labeled wildfire smoke dataset has limited researchers to using small or unbalanced datasets~\cite{ko2012wildfire, jeong2020light}, manually searching for images online~\cite{yin2019recurrent, li2019detection, park2020wildfire, jeong2020light}, or~synthetically generating datasets~\cite{park2020wildfire, zhang2018wildland, yuan2019deep}.

In this work, we propose the following contributions to address the need for a consistent evaluation benchmark for real-world performance: (1) Fire Ignition Library~(FIgLib), a~publicly-available dataset of nearly 25,000 labeled wildfire smoke images captured in Southern California, and~(2) SmokeyNet, a~novel deep-learning-based model using image sequences for real-time wildfire smoke detection. We begin with a review of historical methods of wildfire smoke detection. We then provide details on the FIgLib dataset, the~SmokeyNet architecture, our training procedure, and~our experimental setup. After~a discussion of the results of SmokeyNet's performance against comparable baselines, including human classification performance, we conclude with directions for future~work. 

\section{Related~Work}\label{related}
Before the rise in popularity of deep learning methods, computer vision algorithms leveraging hand-crafted features identified that the visual (e.g., color), spatial, and~temporal (i.e., motion) qualities of smoke are essential for the machine detection of wildfires~\mbox{\cite{ho2009machine, toreyin2009wildfire, genovese2011wildfire, ko2012wildfire}}. More recently, deep learning approaches use a combination of convolutional neural networks (CNNs) \cite{luo2018fire, yin2019recurrent, pundir2019dual, ba2019smokenet, li2019detection, cao2019attention, park2020wildfire, khan2021deepsmoke}, background subtraction~\cite{luo2018fire, cao2019attention, yuan2018detection}, and~object detection methods~\cite{khan2021deepsmoke, zhang2018wildland, li20183d, jeong2020light, jindal2021real} to incorporate visual and spatial features. Long short-term memory (LSTM) networks~\cite{cao2019attention, jeong2020light} or optical flow~\cite{pundir2019dual, yuan2018detection, gupta2021early} methods have been applied to incorporate temporal context from video~sequences. 

Despite many papers reporting above 90\% image classification accuracy for the detection of smoke, the~lack of a large, labeled publicly available wildfire smoke dataset makes it difficult to compare performance between approaches. For~example, Ko~\cite{ko2012wildfire} and Jeong~\cite{jeong2020light} used only 10 and 24 videos in their test sets, respectively, in~which half the videos have smoke and half the videos have no smoke; there are no videos in which a fire starts in the middle of the sequence. With~hundreds of frames per video but so few fire scenes evaluated, the~high accuracies reported may not be representative of real-world performance across different scenarios. Li~\cite{li2019detection} and Park~\cite{park2020wildfire} collected 4595 (36\% positive) and 6354 (22\% positive) wildfire images online, respectively; however, since these images were not from video sequences, the~smoke plumes in the image are more visible, and~likely easier to detect, compared to wildfire smoke initially forming after ignition when seen from a continuous video sequence. Yin~\cite{yin2019recurrent} also manually acquired images online, but~the images represent smoke from a variety of indoor and outdoor scenarios beyond wildland fires. Many works synthetically generate images to overcome the lack of available data; Park~\cite{park2020wildfire} uses generative adversarial networks (GANs), Zhang~\cite{zhang2018wildland} uses live smoke in front of a green screen, and~Yuan~\cite{yuan2019deep} uses computational simulation to generate these images. Given the diversity and challenges of the datasets across these works, it is hard to identify which models are best or choose a particular dataset to use as a benchmark for wildfire smoke~classification.

\citet{govil2020preliminary} is the only work we are aware of that also uses the FIgLib dataset to evaluate wildfire smoke detection performance. They used an InceptionV3 CNN \cite{szegedy2016rethinking} trained from scratch as the primary image classification architecture. Instead of using a sigmoid threshold of 0.5 for the prediction of smoke, as~is common in classification models, a~dynamic threshold was implemented based on the average prediction during the same time of day over the prior three days. This was used to incorporate periodic environmental events (e.g., fog, solar reflection, smog, haze); the data from prior days is not included in the FIgLib dataset, but~is available through the HPWREN Archive. An~analysis of results from this work is discussed in {Section \ref{compareresults}}.

\section{Data}
\unskip
\subsection{FIgLib~Dataset}
\begin{figure}[H]
  
  \includegraphics[scale=0.65]{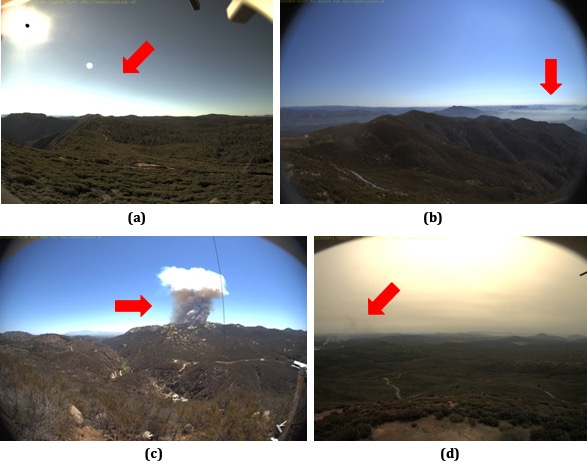}
  \caption{Images from the FIgLib dataset: {(\textbf{a})} no smoke with strong glare, {(\textbf{b})} no smoke with misleading haze, {(\textbf{c})} very apparent wildfire smoke, and~{(\textbf{d})} faint wildfire~smoke. }
  \label{fig-lib-examples}
\end{figure}

The FIgLib dataset addresses the need for a large, labeled publicly-available dataset for wildfire smoke detection. FIgLib reflects sequences of wildland fire images as seen from fixed-view cameras, part of the High Performance Wireless Research and Education Network (HPWREN), on~remote mountain tops in Southern California. As~of December 2021, the~dataset consists of 315 fire sequences from 101 cameras across 30 stations occurring between June 2016 and July 2021. Each sequence typically contains images from 40 min prior to and 40 min following the start of the fire, serving as binary smoke/no-smoke labels for each image, and~are spaced approximately 60 s apart for a total of 81~images per fire sequence. However, 114 fires are missing an average of 6.6 images each; missing images are either at the beginning, end, or~randomly dispersed throughout the sequence. In~total, the~dataset contains 24,800 high-resolution images that are \mbox{1536 $\times$ 2048} or \mbox{2048 $\times$ 3072 pixels} in size, depending on the camera model used. The~ignition detection and view prior to the ignition are enabled by a cluster deployment of cameras, where four 90+ degree views stay consistent for years, covering 360 degrees around a mountaintop. Examples from FIgLib can be seen in {Figure \ref{fig-lib-examples}}, and the full dataset can be accessed at the following link: \url{http://hpwren.ucsd.edu/HPWREN-FIgLib/} (accessed on 16 December~2021). 

\subsection{Data~Preparation}
The number of fires and images in the train, validation, and~test splits of the FIgLib dataset for our machine learning task of wildfire smoke detection are shown in {Table~\ref{dataset-split}}. To~avoid out-of-distribution sequences, we removed fires with black and white images (N = 10), night fires (N = 19), and~fires with questionable presence of smoke (N = 16), including one fire with 180 images and no labels, from~the dataset (3700 images from 45 fires removed in total). In~addition to binary smoke/no-smoke labels for each image, the~smoke in 144~fires has been manually annotated with bounding boxes and contour masks. Since we divide the images into tiles (described in {Section \ref{tiling}}), we can use these annotations to provide smoke/no-smoke labels at a more granular level. Hence, we used images from these 144~annotated fires for training (53.3\% of eligible fires, 11,300~images); the remaining 126 fires (9800 images) were split between the validation and test sets (collectively, the~evaluation sets) such that the number of images in each is roughly equivalent. The~exact list of fires omitted and used for the train, validation, and~test splits can be accessed at the following link: \url{https://gitlab.nrp-nautilus.io/-/snippets/63} (accessed on 16 December~2021).

\begin{table}[H]
  
  \caption{Cross-validation splits of the FIgLib~dataset.}
 \setlength\tabcolsep{46pt}
  \begin{tabular}{lcc}
    \toprule
    \textbf{Model} & \textbf{\# Fires }& \textbf{\# Images} \\
    \midrule
    Train & 144 & 11.3 K \\
    Validation & 64 & 4.9 K \\
    Test & 62 & 4.9 K \\
    \midrule
    Omitted & 45 & 3.7 K \\
    \midrule
    Total & 315 & 24.8 K \\
    \bottomrule
  \end{tabular}
  \label{dataset-split}
\end{table}

While it is uncommon to use such a low percentage of the data for the training set relative to that of the evaluation sets (an 80/10/10 split is more common), our model requires annotations for training and we only had access to 144 annotated fires. We also did not want to omit any fires from the evaluation sets, because~larger evaluation sets provide a more robust representation of how the model would perform in diverse conditions. Splitting the data by fires instead of images ensures that no data related to the test set are in the training set, to obtain a more representative evaluation of the model's real-world performance. Otherwise, the~model would be evaluated on images of fires for which it has already trained on using other frames of the video sequence; the performance of the model in this scenario might be overstated since the model might learn features specific to the particular video sequence that do not extrapolate to completely unseen~scenarios. 

We perform the following transformations during data loading to improve the performance of our model. We first resize the images to the empirically-determined size of 1392 $\times$ 1856 pixels to improve training and inference speed. We also crop the top 352 rows of the image, specifically determined to ensure that only rows well above the horizon are cropped to reduce false positives from clouds for additional performance gains. Resizing and further cropping the height of the images to a final size of 1040 $\times$ 1856 pixels enables us to evenly divide the image into overlapping 224 $\times$ 224 tiles (see {Section \ref{tiling}}). We then randomly apply data augmentations including horizontal flip, vertical crop, color jitter, brightness and contrast jitter, and~blur. Finally, we normalize the images to 0.5 mean and 0.5~standard deviation, as~expected by the deep learning package used (torchvision).

\section{Methods}
\unskip
\subsection{Tiling}\label{tiling}
Our goal is the binary classification of images to determine the presence of smoke as early in the sequence as possible. Training the model with standard CNN techniques by leveraging solely image labels does not provide a sufficient training signal for the model to identify small plumes of smoke within the large images. Object detection models using bounding box and contour mask annotations can better localize the target object using anchors and a regression head~\cite{ren2015faster}; however, these models require precise annotations, which poses a challenge in our scenario given the amorphous and transparent nature of~smoke. 

Consequently, we build upon previous work by tiling the image into 224 $\times$ 224 tiles, overlapping by 20 pixels for a total of 45 tiles~\cite{govil2020preliminary}. We also generate corresponding binary \textit{tile labels}: positive if the number of pixels of smoke in the tile, determined by the filled polygon of the contour mask, is greater than an empirically-determined smoke detection threshold of 250 (0.5\% of the total pixels in the tile). Tile labels provide the entirety of our localized feedback signal; we do not otherwise use the bounding box or contour mask annotations during~training.

One challenge of the dataset is that 1213 (approximately 20\%) of the positive images are missing contour mask annotations. A total of 280 annotations are missing because the smoke is difficult to see, generally occurring at the beginning of the fire sequence or at the end, when the smoke has dissipated. A total of 486 annotations are missing contour masks but have bounding box annotations, generally because the smoke is too small to reasonably outline a fine contour mask. The~remaining 447 missing annotations are randomly spread throughout the~fires. 

For images with bounding box annotations where contour masks are not available, we determined the tile labels by filling the bounding boxes as polygons instead of the contour masks (486 images affected). We attempted other methods to incorporate feedback from images with missing annotations, including using feedback from only image labels (as opposed to both image and tile labels) and copying contour masks from the closest available image in the sequence. However, neither of these methods improved model performance; consequently, we did not train on the remaining positive images with missing annotations (727 images total). For~future work, we aim to resolve these missing labels for more robust training~data.

\subsection{Smokeynet~Architecture}
\begin{figure}[H]
  
  \includegraphics[scale=0.40]{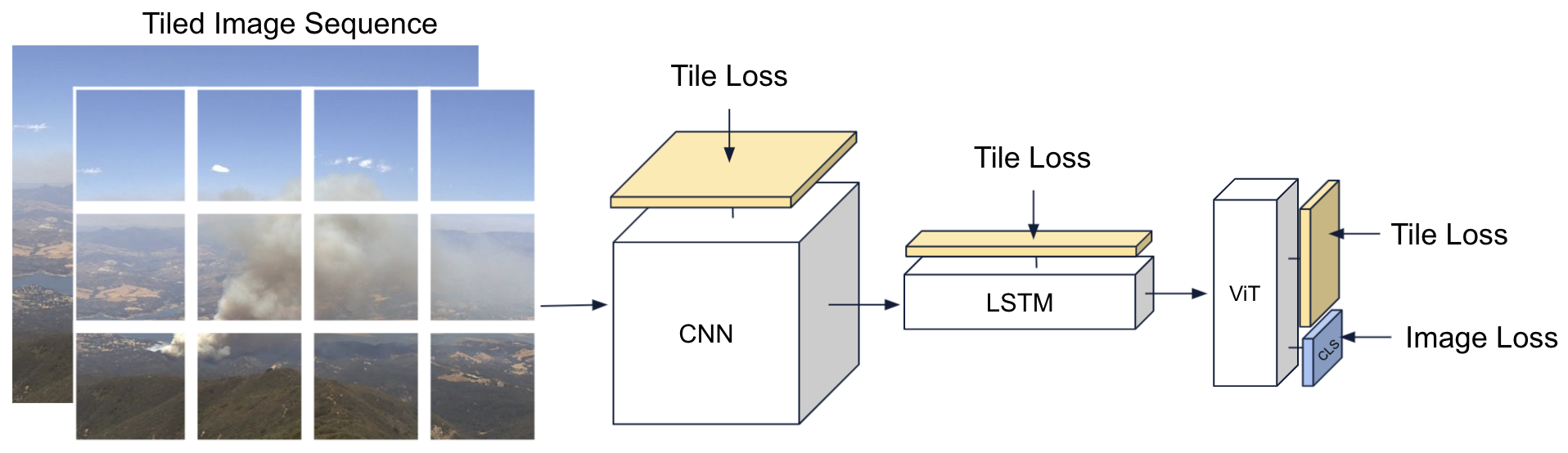}
  \caption{The SmokeyNet architecture takes two frames of the tiled image sequence as input and combines a CNN, LSTM, and~ViT. The~yellow blocks denote ``tile heads'' used for intermediate supervision, while the blue block denotes the ``image head'' used for the final image~prediction. }
  \label{smokeynet}
\end{figure}

The SmokeyNet architecture ({Figure~\ref{smokeynet}}) is a novel spatiotemporal gridded image classification approach for wildfire smoke detection combining three different types of neural networks: a CNN~\cite{krizhevsky2012imagenet}, an~LSTM~\cite{hochreiter1997long}, and~a vision transformer (ViT) \cite{dosovitskiy2020image}. The~input to our model is the tiled raw image and its previous frame in the wildfire video sequence to incorporate the motion of the smoke. A~CNN, pretrained on the ImageNet dataset~\cite{deng2009imagenet}, initially extracts representations of the raw image pixels from each tile of the two frames independently. A~ResNet34, a~lighter-weight version of the popular ResNet50 model, is our preferred choice of CNN backbone~\cite{he2016deep}. Then, an~LSTM combines the temporal information of each tile from the current frame with its counterpart from the previous frame. Finally, all temporally-combined tiles are fed into a ViT, which incorporates spatial information \textit{across} tiles to improve the image~prediction.

\vspace{0.05in}

The outputs of the ViT are spatiotemporal embeddings for each tile, as~well as a CLS token embedding that summarizes representations for the whole image~\cite{dosovitskiy2020image}. The~CLS token embedding is passed to an ``image head'', consisting of three fully-connected layers with ReLU activation with output sizes of 256, 64, and~1, respectively, and~a sigmoid layer with a threshold of 0.5 to generate a single prediction for the whole image. Given the modular nature of each of the components, we can experiment with different approaches to capture spatiotemporal information while still training the model end-to-end. The~full SmokeyNet codebase, including code to run all experiments conducted in this work, can be accessed at the following link: \url{https://gitlab.nrp-nautilus.io/anshumand/pytorch-lightning-smoke-detection} (accessed on 16 December 2021).

\subsection{Loss}
The initial component of our loss applies standard binary cross-entropy (BCE) loss between the outputs of the image head and the ground-truth binary image labels. We can increase the weight of positive examples when calculating this BCE image loss to trade off precision for higher recall. Increasing the positive weight increases the penalty for missing positive examples; while potentially incurring more false positives, the~model will also be able to detect the actual presence of smoke more quickly and accurately, which is of utmost importance.  We use the empirically-determined positive weight of five to achieve more balanced precision and recall and improve the overall accuracy and~F1-score. 

To leverage the localized information provided by the tile labels, we also apply intermediate supervision to each of the model components~\cite{wei2016convolutional}. Since the model's components, the~CNN, LSTM, and~ViT, also produce embeddings on a per-tile basis, we pass each component's embeddings through individual ``tile heads'', consisting of three fully-connected layers with ReLU activation with output sizes of 256, 64, and~1, respectively, and~a sigmoid layer to generate predictions for each \textit{tile}. We then apply BCE loss between the outputs of the tile heads and the binary tile labels. To~address the class imbalance in which negative tiles occur more frequently than positive tiles, we weight positive examples by 40, the~ratio of negative tiles to positive~tiles.

If $I$ is the total number of tiles, the~overall training loss can be summarized as
\begin{center}
    $loss = \text{BCE}^{image}  +  \sum_i^I \{\text{BCE}_i^{\text{CNN}}  +  \text{BCE}_i^{\text{LSTM}}  +  \text{BCE}_i^{\text{ViT}}\}$
\end{center}

Since we have tile labels for only the training data, we define our validation loss as the average number of \textit{image} prediction errors and use this validation loss for early stopping. The~BCE loss equations are elaborated upon in {Appendix \ref{bceloss}}.

\section{Experiments}
\unskip
\subsection{Comparable~Baselines}
We experiment with alternate CNN backbones to the ResNet34, including a MobileNetV3Large (denoted ``MobileNet'') \cite{howard2019searching}, MobileNet with a feature pyramid network (FPN) \cite{lin2017feature} to better incorporate spatial scales, EfficientNet-B0~\cite{tan2019efficientnet}, and~Data Efficient Image Transformer (DeiT-Tiny) \cite{touvron2021training}. Using the ResNet34 as the backbone, we also try inputting three frames (i.e., two additional frames of temporal context) instead of two and conduct an ablation study by removing different parts of the model to evaluate each component's benefits. We then experiment with different architectures that can capture the temporal information from sequential frames, including replacing the LSTM with a transformer~\cite{vaswani2017attention}; using a CNN and a ResNet18-3D CNN to replace both the LSTM + ViT~\cite{tran2018closer}; and incorporating motion information using MOG2, a~Gaussian-mixture-based background removal method~\cite{zivkovic2004improved, zivkovic2006efficient}, as~an additional channel of input. Finally, we compare the model's performance to three baseline architectures: ResNet50, the~standard for image classification models~\cite{he2016deep}; faster-RCNN, a~standard object detection model~\cite{ren2015faster}; and mask-RCNN, an~image segmentation model leveraging both contour masks as well as bounding boxes for training signal~\cite{he2017mask}. 

For baseline models that do not use a ViT as the last architectural component (e.g., ResNet34 + LSTM, ResNet50, ResNet34 + ResNet18-3D, etc.), there is no CLS token embedding that summarizes representations for the whole image that we can use for our image prediction. Consequently, we determine the overall image prediction by passing the model's tile predictions into a single fully connected layer with sigmoid activation, outputting a single prediction for the image. We also experimented with the simple decision rule that if the prediction for any tile is positive for smoke, the~full image is also classified as positive; however, this resulted in worse performance. Image predictions for object detection models (e.g., faster-RCNN, mask-RCNN) were determined as positive if the model predicted any bounding box with a confidence score above the empirically-determined threshold of 0.5 (\{0, 0.2, 0.4, 0.5, 0.6\} all tested). Additional model implementation details for alternative architectures are described in {Appendix \ref{trainingdetails}}.

\subsection{Training~Details}
Hyperparameter tuning was performed sequentially, with~the best result from one set of experiments used in subsequent experiments, in~the following order: learning rate (\{1e-2, \textbf{1e-3}, 1e-4\}), weight decay (\{1e-4, \textbf{1e-3}\}), image resizing (\{100\%, \textbf{90\%}, 80\%, 50\%\} of 1536 $\times$ 2048), smoke detection threshold to determine tile labels (\{0, 10, 100, \textbf{250}\} pixels per tile), dropout (\{\textbf{0}, 0.1\}), and~image BCE loss positive weight (\{1, 2, \textbf{5}, 10\}). Final models were trained using an SGD optimizer with learning rate 0.001, weight decay 0.001, and~no dropout. The~batch size used was the larger of 2 or 4, depending on which would fit into GPU memory, and~gradient batches were accumulated such that the effective batch size was 32. Models were trained for 25 epochs using a single NVIDIA 2080Ti GPU; the model with the lowest validation loss was used for evaluation on the test~set. 

\subsection{Evaluation~Metrics}
For each experiment, we report the following evaluation metrics typical for binary classification problems:
\begin{center}
$Accuracy = \frac{TP + TN}{TP + TN + FP + FN}$
$Precision = \frac{TP}{TP + FP}$
$Recall = \frac{TP}{TP + FN}$
$F1 = \frac{2*Precision*Recall}{Precision + Recall}$
\end{center}
derived from true positives (\emph{TP}), false positives (\emph{FP}), true negatives (\emph{TN}), and~false negatives (\emph{FN}) calculated between the model predictions and the ground truth labels. We also report the average time-to-detection, calculated as the number of minutes until the model correctly predicts the first positive frame of a wildfire sequence, averaged over all fires. We include the number of parameters (in millions) and inference time (ms/image) of each model, which should be minimized for deployment to edge~devices.

\section{Results}
\unskip
\subsection{Experimental~Results}
{Table~\ref{results}} reports test evaluation performance for each of the experimental architectures with two frames of input (unless otherwise stated) as an average over five runs. The~SmokeyNet architecture with a ResNet34 backbone and two frames of input achieves an image accuracy of 83.49\% and F1-score of 82.59\% while delivering on the objectives of high precision (89.84\%), high recall (76.45\%), fast performance (51.6ms/image), and~low average time to detection (3.12 min). One additional frame of input (ResNet34 + LSTM + ViT (3 frames)) only marginally improves performance at the cost of a 55.6\% increase in inference time. Large backbones, such as the ResNet34 or EfficientNet-B0, trade off model size and inference time for better accuracy compared to smaller backbones, such as the MobileNet or~MobileNetFPN.

From the ablation study, we observe that the standalone CNN or CNN + LSTM models perform poorly at the task. Adding the ViT to the CNN significantly improves performance with little impact to inference speed. All three alternate architectures to incorporate temporal information perform slightly worse than SmokeyNet; however, the~ResNet34 + ResNet18-3D architecture provides another viable alternative if prioritizing model size. Comparing MobileNet + LSTM + ViT to MobileNet + LSTM + ViT(MOG2), we see that the addition of background subtraction improves performance by almost 2\%, mainly by improving recall by over 3\%, while maintaining high precision. This comes at the cost of model size and inference time, but~it is worth exploring adding MOG2 as another SmokeyNet variant in the~future. 

\begin{table}[H]
  \small
  \caption{Accuracy (A), F1, precision (P), recall (R), and average time-to-detection (TTD) evaluation metrics on the test set. Best results are \textbf{bolded}.}
  \begin{tabular}{lccccccc}
    \toprule
\multirow{2}{*}{\textbf{Model}}   &\textbf{Params} &\textbf{Time}  & \multirow{2}{*}{\textbf{A}} & \multirow{2}{*}{\textbf{F1}} & \multirow{2}{*}{\textbf{P}} & \multirow{2}{*}{\textbf{R}} &{\textbf{TTD}}  \\

&\textbf{(M)}&\textbf{(ms/it)}&&&&&\textbf{(mins)}\\
    \midrule
    \textbf{Variants of SmokeyNet:}  &&&&&& \\
    ResNet34  +  LSTM  +  ViT & 56.9 & 51.6 & 83.49 & 82.59 & 89.84 & 76.45 & 3.12 \\
    ResNet34 + LSTM + ViT (3 frames)  & 56.9 & 80.3 & \textbf{83.62} & \textbf{82.83} & \textbf{90.85} & 76.11 & 2.94 \\
    MobileNet  +  LSTM  +  ViT        & 36.6 & 28.3 & 81.79 & 80.71 & 88.34 & 74.31 & 3.92 \\
    MobileNetFPN  +  LSTM  +  ViT     & 40.4 & 32.5 & 80.58 & 80.68 & 82.36 & 79.12 & 2.43 \\
    EfficientNetB0  +  LSTM  +  ViT   & 52.3 & 67.9 & 82.55 & 81.68 & 88.45 & 75.89 & 3.56 \\
    TinyDeiT  +  LSTM  +  ViT         & 22.9 & 45.6 & 79.74 & 79.01 & 84.25 & 74.44 & 3.61  \\
    \midrule
    ResNet34 (1 frame)            & 22.3 & 29.7 & 79.40 & 78.90 & 81.62 & 76.58 & 2.81 \\
    ResNet34  +  LSTM               & 38.9 & 53.3 & 79.35 & 79.21 & 82.00 & 76.74 & 2.64 \\
    ResNet34  +  ViT (1 frame)      & 40.3 & 30.8 & 82.53 & 81.30 & 88.58 & 75.19 & 2.95 \\
    \midrule
    ResNet34  +  Transformer  +  ViT  & 58.9 & 50.6 & 82.74 & 81.39 & 90.69 & 73.95 & 3.18 \\
    ResNet34  +  ResNet18-3D        & 38.0 & 57.5 & 83.10 & 82.26 & 88.91 & 76.65 & 2.87 \\
    MobileNet + LSTM + ViT (MOG2)  & 57.3 & 55.5 & 83.12 & 82.65 & 88.79 & 77.43 & 3.72 \\
    \midrule
    ResNet50 (1 frame)            & 26.1 & 50.4 & 68.51 & 74.30 & 63.35 & \textbf{89.89} & \textbf{1.01} \\
    FasterRCNN (1 frame)          & 41.3 & 55.6 & 71.56 & 66.92 & 81.34 & 56.88 & 5.01 \\
    MaskRCNN (1 frame)            & 43.9 & 56.9 & 73.24 & 69.94 & 81.08 & 61.51 & 4.18 \\
    \bottomrule
  \end{tabular}
  \label{results}
\end{table}

The SmokeyNet architecture clearly outperforms standard image classification, object detection, and~image segmentation baselines. All models have acceptable time to detection---most within 3--4 min. While the ResNet50 is clearly the fastest to detect a fire, it only accomplishes this by generating an unacceptable amount of false~positives.

\subsection{Performance~Visualization}
\begin{figure}[H]
  
  \includegraphics[scale=0.5]{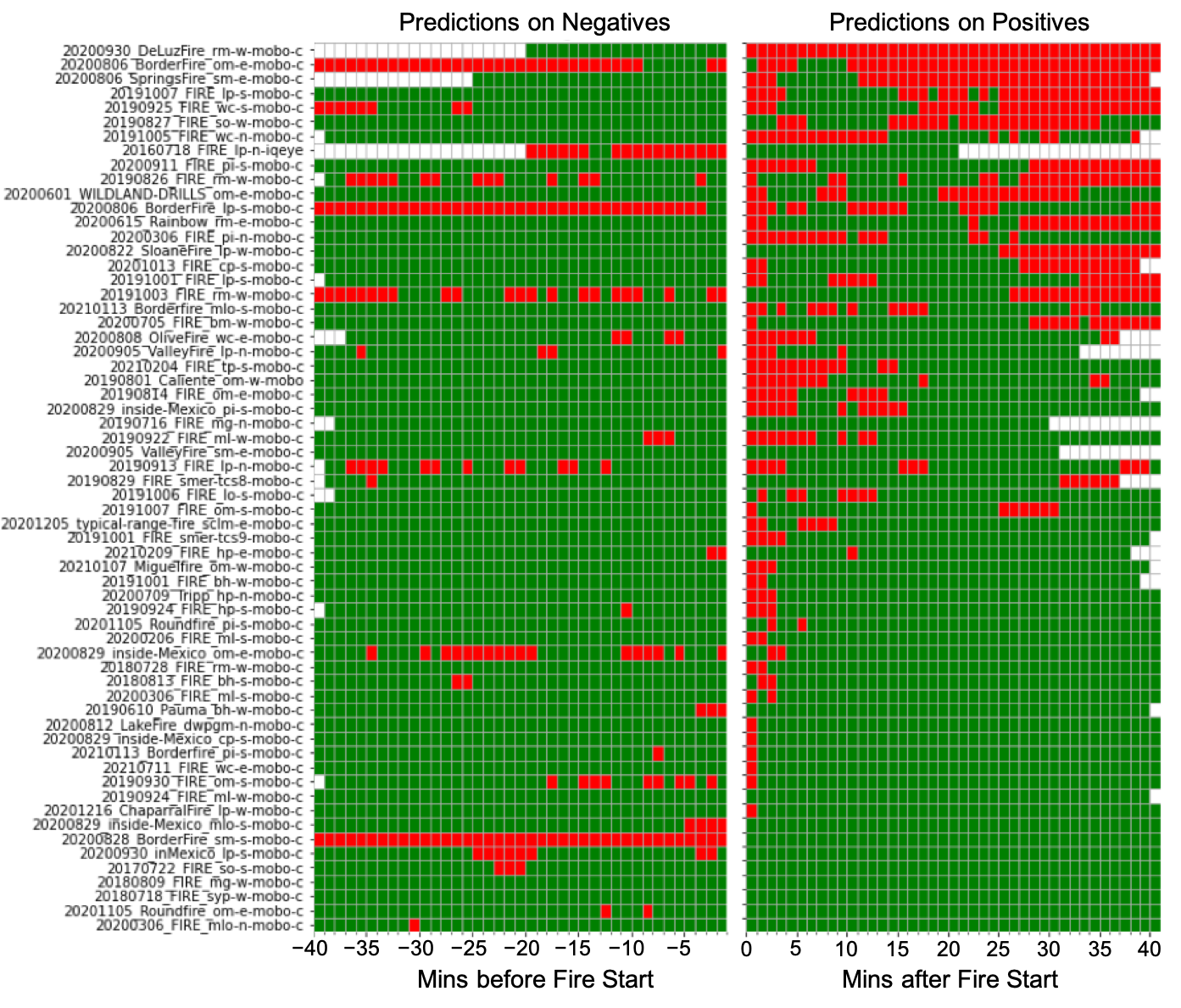}
  \caption{SmokeyNet's performance per fire on both negative and positive images. Green denotes a correct prediction; red denotes an incorrect prediction; white denotes images missing from the sequence. Hence, red on the left are false positives, red on the right are misses. Common misses include faint smoke occurring at the start of the fire or dissipating smoke at the end of the fire sequence. Common false positives include low-altitude clouds and~haze.} 
  \label{test-set-performance}
\end{figure}

{Figure~\ref{test-set-performance}} (on the following page) further visualizes SmokeyNet's performance on images from the test set. Additionally, a~video of the model's performance per image can be viewed at the following link: \url{https://youtu.be/cvXQJao3m1k} (accessed on 16 December 2021). The~model performs well in a variety of real-world scenarios, correctly identifying apparent smoke plumes while avoiding clouds and haze. However, the~model still makes systematic misclassifications of low-altitude clouds as false positives (\texttt{20200806\_BorderFire\_om-e-mobo-c}, \texttt{20200828\_BorderFire\_sm-s-mobo-c},
\texttt{20200806\_B-}
\texttt{orderFire\_lp-s-mobo-c}). The~model completely missed the \texttt{20200930\_}
\texttt{DeLuzFire\_rm-w-}
\texttt{mobo-c} fire (top row of {Figure~\ref{test-set-performance}}) because the smoke occurs directly behind a transmission~tower.

\subsection{Human Performance~Baseline}
Due to the lack of suitable benchmarks for performance, we additionally measured human performance of smoke classification on the FIgLib dataset. Participants were three lab members experienced in classifying images for the presence of wildfire smoke. For~the experimental setup, one image from each of the 62 fires from the test set was randomly selected for prediction. Participants were presented with the images for prediction, each preceded by the previous frame of the image sequence, to~replicate the temporal context our machine learning model receives and a real-time inference scenario. The~participants then recorded if they believed wildfire smoke was present in the~image.

The experimental setup proves challenging due to differences from how the ground truth labels were generated. The~ground truth labels were created by a human expert who had full temporal context, including access to all past and \textit{future} frames of a video sequence, to~precisely determine the first frame in which smoke is visible. Consequently, it is difficult to correctly identify positive images early in the fire sequence when the smoke is faint or small with only a single preceding frame and no additional temporal context to see how the smoke grows over~time.

The three participants achieved an average accuracy of 78.5\% ($\sigma$ = 1.52\%), F1-score of 82.8\% ($\sigma$ = 0.73\%), precision of 93.5\% ($\sigma$ = 4.66\%), and~recall of 74.4\% ($\sigma$ = 1.90\%). The~low accuracy and recall signify the false negatives from missing positive images early in the fire sequence. SmokeyNet achieved a higher accuracy and similar F1-score compared to human performance. The~model also had higher recall at the expense of lower precision. If~the false positives derived from low-altitude clouds could be corrected, SmokeyNet would achieve 85.8\% accuracy and 94.8\% precision, matching the precision and further surpassing the accuracy of human performance. Future work should focus on this~issue.

\section{Discussion}
\unskip
\subsection{Innovations in the SmokeyNet~Architecture}
SmokeyNet establishes a strong baseline for performance on the FIgLib dataset for the task of wildfire smoke detection. With~a finely-tuned custom architecture, SmokeyNet improves upon works that leverage standard out-of-the-box models for image classification or smoke detection~\cite{govil2020preliminary, khan2021deepsmoke, jindal2021real}. As~the first transformer-based architecture for wildfire smoke detection that we are aware of, SmokeyNet should see improved performance over CNNs when training on even larger datasets~\cite{vaswani2017attention}. SmokeyNet also includes features that enable quick experimentation for further improvement. Modularity allows the user to easily swap out components of the SmokeyNet architecture to test alternate backbones or temporal aggregators. Intermediate supervision mitigates the likelihood of unstable training even with very large models~\cite{wei2016convolutional}. Lastly, the~tiling of input images enables SmokeyNet to adapt to input images of any~size.

\subsection{Comparison of FIgLib to Previous~Datasets}
The structure of the FIgLib dataset enables a more robust evaluation of machine learning models for wildfire smoke detection in real-world scenarios than datasets in previous work. Since FIgLib depicts fires from only wildland areas, models trained on FIgLib for wildfire smoke detection avoid a data distribution shift that would occur when training models on datasets with both indoor and outdoor scenes~\cite{yin2019recurrent}. FIgLib's use of video sequences, as~opposed to the static images used in~\citet{li2019detection} and~\citet{park2020wildfire}, allows models to leverage temporal data to improve predictions. FIgLib includes frames in sequences before and after the initial fire ignition, encouraging models to learn to detect the ignition itself, when smoke plumes are small. This is of great practical importance, as~fast time-to-detection can prevent fires from getting out of control. These features contrast with the datasets used in Ko~et~al.~\cite{ko2012wildfire} and Jeong~et~al.~\cite{jeong2020light}, in~which the videos contain either all smoke or no smoke, making it difficult for a model to learn to detect fire ignition. Furthermore, FIgLib represents an order of magnitude more data, 315 fire sequences from 101 cameras, versus the 10 and 24 videos used in previous work~\cite{ko2012wildfire,jeong2020light}. This variety ensures models are trained and evaluated in a number of different scenarios to avoid overfitting. Going forward, we plan to continue adding more recent fires to the FIgLib dataset, expanding the HPWREN camera network, and, upon~further validation, releasing the human-annotated bounding boxes and contour masks to assist model~training.

It is important to note that the FIgLib dataset contains roughly 50\% positive examples and 50\% negative examples representing frames 40 min before and after the start of each fire. While a balanced dataset makes it easy to train a machine learning model for binary classification, one limitation is that this ratio is not representative of real-world scenarios in which positive examples of visible smoke are much more rare than negative, no-smoke examples. This may induce models to incur a higher false positive rate on real-world data. To~mitigate this, models can leverage unlabeled data from the HPWREN Archive (\url{http://c1.hpwren.ucsd.edu/archive/}, accessed on 16 December 2021) in combination with FIgLib to obtain a more representative sample of positive and negative examples to fine-tune the model for real-world~deployment.

\subsection{Comparison of SmokeyNet Performance to Previous~Work}\label{compareresults}
It is impossible to make a head-to-head performance comparison with previous work due to differences in the datasets used and evaluation metrics reported, as~discussed in {Section \ref{related}}. \citet{govil2020preliminary}, the~only work to also use the FIgLib dataset, reported a test accuracy of 0.91 and F1-score of 0.89, higher than SmokeyNet's accuracy of 0.83 and F1-score of 0.82. However, their test set consisted of only 250 hand-selected images relative to the training set of about 8500 images. The~test set also contained images from the same video sequences of fires used in the training set, with~only 10 min (i.e., 10 frames) of separation between them. This procedure suggests that the test data may only be 10 min earlier or later than the training data, and~therefore may overstate performance. Completely independent test data is an important methodological requirement in machine learning to measure true generalization performance. The~authors also reported results from field-testing the model on 65 HPWREN cameras over a period of nine days in October 2019. After~suppressing repeat detections in a one-hour timespan, only 21\% of notifications showed smoke from real fires (i.e., a~79\% false positive rate) and there was no report of how many actual fires were~missed. 

The standard in machine learning when comparing two models is that they use the same training, validation, and~test sets. Consequently, despite SmokeyNet's numerically lower performance on the evaluation set, it is difficult to make a direct comparison with the performance of the model in \citet{govil2020preliminary}. Given the much larger training and test sets used, we believe SmokeyNet is more robust to unseen scenarios and more representative of real-world performance. Further experimentation should compare SmokeyNet to other architectures~\cite{govil2020preliminary, gupta2021early, li20183d, pundir2019dual, yin2019recurrent, li2019detection} using consistent training and evaluation splits from FIgLib as the basis for fair comparison. Additionally, it is important to follow the example of \mbox{\citet{govil2020preliminary}} and test the model on unseen, real-world~data.

\subsection{Planned Future~Work}
For future work, we will continue improving the performance of SmokeyNet by reducing false positives in difficult scenarios such as low-altitude clouds and haze, a~current limitation of the model. One method would be to take false positives from the holdout set and incorporate them into the training set as in \citet{govil2020preliminary}; however, in~order to maintain the integrity of the holdout set, we would have to split our holdout set in two and leave one untouched when following this procedure. Advanced preprocessing techniques can extract candidate smoke patches before inputting the image into the model by using ViBE background removal~\cite{barnich2010vibe}, YUV color space segmentation~\cite{li2019detection}, or~a dark channel prior~\cite{gupta2021early, luo2018fire} for potential performance gains. Since fires frequently occur in similar locations and conditions during specific months of the year, incorporating fire location, date, and~weather data from historical fire records can further improve~predictions. 

We also plan to utilize the large amount of unlabeled data from the HPWREN Archive. We can leverage SmokeyNet's performance to make predictions on unlabeled data, which can then be validated by human inspection. These validated predictions can then be used to quickly expand the amount of labeled training data. The~unlabeled data can also be used in the self-supervised representation learning setting, using models such as DINO~\cite{caron2021emerging}. This approach allows the model to learn hidden representations from large amounts of unlabeled data to better perform in a classification task with limited labeled data. Since the vast majority of the unlabeled images will not contain fires, we can alternatively use generative adversarial networks (GANs) to synthetically generate smoke in these images to produce positive examples for additional training data as in \citet{park2020wildfire}. By~effectively increasing the size of our training data and extracting better representations from the data, these approaches enable our model to make better predictions of the presence of~smoke.

Lastly, we plan future work to reduce the model size for better compatibility with edge devices, another limitation of SmokeyNet, without~sacrificing prediction performance. We can use pruning~\cite{pan2021fourier} to eliminate sparsely-used model weights to decrease model size. We can also use distillation to train a smaller deep network to perform similarly to our current model; this strategy will result in a model more suitable for edge computing~\cite{hinton2015distilling, jeong2020light}.

\section{Conclusions}
FIgLib addresses the need for a large, labeled dataset that incorporates a variety of scenarios for the early detection of wildfire smoke. SmokeyNet provides a strong baseline for automated wildfire smoke detection. Since SmokeyNet is trained and evaluated on the publicly-available FIgLib dataset, its performance can be easily compared to subsequent approaches. We ultimately hope our contributions enable further research and development of automated notifications for wildfire smoke~detection.

\vspace{6pt} 



\authorcontributions{Conceptualization, M.H.N., Y.P., A.D. and G.W.C.; methodology, A.D., Y.P., M.H.N. and G.W.C.; software, A.D. and Y.P.; validation, A.D., Y.P., M.H.N. and G.W.C.; formal analysis, A.D. and Y.P.; investigation, A.D. and Y.P.; resources, H.-W.B., F.V. and I.A.; data curation, H.-W.B., F.V. and I.P.; writing---original draft preparation, A.D.; writing---review and editing, A.D., M.H.N., G.W.C., I.A., H.-W.B., F.V. and Y.P.; visualization, A.D.; supervision, M.H.N. and G.W.C.; project administration, M.H.N.; funding acquisition, I.A. All authors have read and agreed to the published version of the~manuscript.}

\funding{This research was funded in part by NSF grant numbers 1730158, 2100237, 2120019 for Cognitive Hardware and Software Ecosystem Community Infrastructure (CHASE-CI) and 1331615, 2040676 and 1935984 for WIFIRE, WIFIRE Commons, and~SAGE.}


\institutionalreview{Not applicable.}

\informedconsent{\textls[-5]{Informed consent was obtained from all subjects involved in the study.}}

\dataavailability{Publicly available datasets were analyzed in this study. This data can be found here: \url{http://hpwren.ucsd.edu/HPWREN-FIgLib/}, (accessed on 16 December 2021).} 

\acknowledgments{The authors would like to thank Brian Norton for sharing his invaluable expertise on wildfire management; Stephen Jarrell, Duolan Quyang, Atman Patel, and~Ulyana Tkachenko for their collaboration and insights; Scott Ehling and his team of labellers at Argonne National Laboratory for contributing to dataset annotations; and the UC San Diego CHASE-CI team for computer~support.}

\conflictsofinterest{The authors declare no conflict of~interest.}





\appendixtitles{yes} 
\appendixstart
\appendix

\section{Summary of~Hyperlinks}\label{hyperlinks}
The following hyperlinks were referenced in the body of the work and are listed again here for convenience (in order of appearance):
\begin{itemize}
    \item FIgLib Dataset: \url{http://hpwren.ucsd.edu/HPWREN-FIgLib/}, (accessed on 16 December 2021)
    \item Cross-Validation Splits of FIgLib Dataset: \url{https://gitlab.nrp-nautilus.io/-/snippets/63}, (accessed on 16 December 2021)
    \item Codebase for SmokeyNet \& Related Experiments: \url{https://gitlab.nrp-nautilus.io/anshumand/pytorch-lightning-smoke-detection}, (accessed on 16 December 2021)
    \item Visualization of SmokeyNet Performance: \url{https://youtu.be/cvXQJao3m1k}, (accessed on 16 December 2021)
    \item HPWREN Archive: \url{http://c1.hpwren.ucsd.edu/archive/}, (accessed on 16 December 2021)
\end{itemize}

\section{Binary Cross-Entropy Loss~Equations}\label{bceloss}
Standard binary cross-entropy (BCE) loss for the two-class case is summarized in {Equation \eqref{e1}}, in~which $N$ is the number of examples, $y$ is the ground truth label, and~$p$ is the model prediction:
\begin{equation}\label{e1}
    \text{BCE} = -\frac{1}{N}\sum_{n=1}^N{(y_n\log(p_n)  +  (1 - y_n)\log(1 - p_n))}
\end{equation}

The total loss used in SmokeyNet is summarized in {Equation \eqref{e2}}, in~which $I$ is the total number of tiles:
\begin{equation}\label{e2}
    loss = \text{BCE}^{image}  +  \sum_i^I \{\text{BCE}_i^{\text{CNN}}  +  \text{BCE}_i^{\text{LSTM}}  +  \text{BCE}_i^{\text{ViT}}\}
\end{equation}

Modified from the standard BCE loss equation, {Equations \eqref{e3}--\eqref{e5}} elaborate upon the tile losses of the CNN, LSTM, and~ViT, in~which $y^{tile}$ are the tile labels, $p^{\text{CNN}}$ are the outputs of the CNN, $p^{\text{LSTM}}$ are the outputs of the LSTM, and~$p^{\text{ViT}}$ are the outputs of the ViT. Note that the weight of the positive examples is 40 to address class imbalance:
\begin{equation}\label{e3}
    \text{BCE}^{\text{CNN}} = -\frac{1}{N}\sum_{n=1}^N{(40*y_n^{tile}\log(p_n^{\text{CNN}})  +  (1 - y_n^{tile})\log(1 - p_n^{\text{CNN}}))}
\end{equation}
\begin{equation}\label{e4}
    \text{BCE}^{\text{LSTM}} = -\frac{1}{N}\sum_{n=1}^N{(40*y_n^{tile}\log(p_n^{\text{LSTM}})  +  (1 - y_n^{tile})\log(1 - p_n^{\text{LSTM}}))}
\end{equation}
\begin{equation}\label{e5}
    \text{BCE}^{\text{ViT}} = -\frac{1}{N}\sum_{n=1}^N{(40*y_n^{tile}\log(p_n^{\text{ViT}})  +  (1 - y_n^{tile})\log(1 - p_n^{\text{ViT}}))}
\end{equation}

Finally, {Equation \eqref{e6}} expands upon the overall image loss, in~which $y^{image}$ are the image labels and $p^{image}$ are the outputs of the image head corresponding to the final outputs of the model. Note that the positive weight is only 5 in this case:
\begin{equation}\label{e6}
    \text{BCE}^{image} = -\frac{1}{N}\sum_{n=1}^N{(5*y_n^{image}\log(p_n^{image})  +  (1 - y_n^{image})\log(1 - p_n^{image}))}
\end{equation}

\section{Experimental Architecture~Details}\label{trainingdetails}
 A feature pyramid network (FPN) is a computer vision architecture that better recognizes spatial scales by incorporating information from multiple layers of the CNN backbone~\cite{lin2017feature}. Instead of producing a single 960-dimensional embedding like a standard MobileNetV3Large~\cite{howard2019searching}, the~MobileNetFPN outputs 3 layers of 256-channel feature maps per tile sized 7 $\times$ 7, 7 $\times$ 7, and~4 $\times$ 4 respectively. We downsample each feature map through two convolutional layers of kernel size 1 and flatten the feature maps such that the total number of features per map is 784. We then concatenate all the feature maps and further downsample the concatenated features to 960 to match the embedding size of the standard MobileNetV3Large. These are the final embeddings that are passed onto the next component of the SmokeyNet architecture, the~LSTM.
 
 To incorporate MOG2 background removal as an additional input channel, we first take two sequential frames of the raw wildfire smoke video sequence and apply MOG2 background removal; this generates a single channel of dimensions equivalent to the raw image inputs. While two sequential frames of the tiled raw images are input into the CNN and LSTM of the SmokeyNet architecture, two sequential frames of the MOG2 channel are  passed through a separate CNN and a separate LSTM. The~embeddings resulting from both LSTMs, one for the raw image and one for the MOG2 channel, are then concatenated and passed through a single linear layer downsampling the feature maps by half before being passed onto the next component of the standard SmokeyNet architecture, the~ViT.


\begin{adjustwidth}{-\extralength}{0cm}
\reftitle{References}


\end{adjustwidth}

%



\end{document}